\documentclass[runningheads]{llncs}

\usepackage{accv}

\usepackage{accvabbrv}
\usepackage[nolist]{acronym}
\usepackage[T1]{fontenc}
\usepackage{graphicx}
\usepackage{amsmath,amssymb}
\usepackage{booktabs}
\usepackage{multirow}
\usepackage{xcolor}
\usepackage[expansion=false]{microtype}
\usepackage{enumitem}

\usepackage[accsupp]{axessibility}  

\usepackage[pagebackref,breaklinks,colorlinks,citecolor=accvblue,bookmarks=true]{hyperref}

\usepackage{orcidlink}

\newcommand{\best}[1]{\textbf{#1}}
\newcommand{\second}[1]{\underline{#1}}
\newcommand{\loglam}{\log\hat{\lambda}}

\acrodef{GNN}{Graph Neural Network}

\begin{document}

\title{Rate as a Feature, Not a Nuisance:\protect\\
       Curriculum FiLM Training for Event-Based\\
       Graph Neural Networks}
\titlerunning{Curriculum-FiLM Event GNN}
\author{Kavyansh Tyagi\inst{1}\orcidlink{0009-0008-5830-2999} \and Andreas Ziegler\inst{2}\orcidlink{0009-0004-4477-2042}}
\authorrunning{K.~Tyagi and A.~Ziegler}
\institute{National Institute of Technology Kurukshetra, Kurukshetra, India\\
\email{123109026@nitkkr.ac.in}
\and
University of Toronto, Toronto, Canada\\
\email{andreas.ziegler@robotics.utias.utoronto.ca}}
\maketitle

\begin{abstract}
An event camera emits events at a rate given by the scene motion, object texture, illumination, its own contrast thresholds, and other settings.
Across the N-Caltech101 dataset, recorded by a single event camera, the event rate varies by roughly $26\times$.
Standard event-based \ac{GNN} pipelines do not make use of the variable event rate.
They either normalize each event's timestamp into a fixed window or train the network to be invariant to the event rate.
We argue that both choices lose information that could instead be used to improve the performance of a pipeline.
Per-recording normalization is non-invertible, incurring about 6\,pp (percentage points) of natural-rate accuracy loss in our setting.
And invariance training trades accuracy for robustness to shifts in the event rate.
We instead take a different path and condition on the event rate directly.
We add a per-recording log-rate scalar that modulates every stage of a PointNet++ backbone through Feature-wise Linear Modulation (FiLM), initialized to identity, so that training begins at the unmodified baseline.
A two-stage curriculum (natural event rates first, then rate-augmented fine-tuning) reaches 67.2\% on N-Caltech101 (+6.3\,pp) and 82.5\% on DVS128 Gesture (+14.0\,pp) with an additional ${\sim}0.2$M parameters.
A matched-capacity control experiment shows that the gain is not from the increase in model capacity but from the rate conditioning.
We will make the implementation publicly available upon acceptance of this work.
\end{abstract}

\acresetall

\section{Introduction}
\label{sec:intro}

Event cameras report brightness changes at each pixel as they happen, rather than capturing full images at fixed intervals~\cite{lichtsteiner2008}.
A pixel emits a signal, called an \textit{event}, only when its log-intensity changes by more than a fixed threshold.
Therefore, the resulting stream is sparse, asynchronous, and has a temporal resolution on the order of $\mu$s.
This makes event cameras well suited for scenes with fast motion as well as poor and changing illumination, conditions in which conventional frame-based cameras either blur or saturate.
These properties have motivated the use of event cameras in robotics, automotive perception, and other settings with fast movement, unreliable lighting, and limited power budgets.

Because individual events arrive at no fixed rate and can be unevenly distributed in space, an event stream is naturally closer to a point cloud than to a sequence of images.
Representing events this way, as a set of points with spatial coordinates, a timestamp, and a polarity, connected into a graph over local neighborhoods, preserves this irregularity instead of forcing it into a dense grid~\cite{bi2020graph,schaefer2022aegnn}.
PointNet++~\cite{qi2017pointnet++} is a natural backbone for this representation: it groups points into local neighborhoods at increasing spatial scales and is designed to handle point sets whose local density varies across the input, which is the case for event streams.

That local density of events is given jointly by how fast objects move, how textured they are, how bright the scene is, and how the sensor itself is configured.
As a result, the overall rate at which a scene produces events, the number of events per unit time within a recording, can vary enormously even within a single dataset captured with a single camera:
\eg{}, on N-Caltech101 we measure a spread of roughly $26\times$ between the slowest and fastest recordings in the test set.
Recordings of different classes tend to produce systematically different event rates, so the event rate can carry information about what is in the scene.
A natural concern is that the model could exploit event rate as a shortcut for predicting the class label. We therefore quantify the statistical dependence between event rate and class and find that they are only weakly coupled.
This weak coupling makes a rate-only explanation insufficient for the
observed gains, supporting the interpretation that the model learns
rate-conditional feature representations rather than relying on event rate as a direct class cue.

Existing neural network architectures for event-based data do not make use of the event rate.
Some approaches rescale each recording's timestamps into a fixed temporal window before training, thereby removing the original event rate by construction~\cite{schaefer2022aegnn}.
Others apply temporal rescaling during training to encourage invariance to event-rate variations~\cite{bi2020graph,zhu2019eventgan}.
We argue that both choices have drawbacks.
The first discards information that is useful for prediction because the rescaling is not invertible, and the second buys robustness to synthetic time shifts that no natural recording produces, at the cost of accuracy on natural recordings.
We propose to keep the event rate and use it directly, rather than removing it or training the network to ignore it.
A single scalar value, summarizing each recording's event rate, is given to the network as an explicit input:
it modulates the features at every stage of a PointNet++ backbone through Feature-wise Linear Modulation (FiLM)~\cite{perez2018film}.

A model trained this way on event data with natural event rates performs well under realistic conditions but degrades sharply when a recording's event rate falls far outside the training distribution.
Training with randomized rate rescaling restores this robustness, but at the cost of the accuracy gained from rate conditioning.
We resolve this with a two-stage procedure: we first train with rate conditioning on natural rates only, then continue training the same network with randomized rate rescaling.
This approach retains most of the accuracy gained in the first stage while regaining robustness to shifts in the event rate.

To test whether the gain is architecture-specific, we apply the same recipe unmodified to a state-space backbone and check whether the same ordering holds.

In summary, this paper makes the following contributions:
\begin{enumerate}
    \item \textbf{Method:} Per-layer rate conditioning via a log-rate scalar input to
    FiLM layers, with matched-parameter and placement ablations showing the gain
    comes from the rate signal rather than added capacity, and a shortcut-classifier
    control ruling out rate-label leakage.

    \item \textbf{Curriculum:} A two-stage warm-start training recipe that improves
    natural-rate accuracy across datasets, with a test-time extension for extreme
    rate shifts.

    \item \textbf{Cross-architecture generalization:} The same recipe, applied
    unmodified to a state-space backbone (EventMamba) structurally unlike
    PointNet++, reproduces the same ordering, evidence the gain is a property
    of the recipe, not of one architecture.
\end{enumerate}

\section{Related Work}
\label{sec:related}

The field has explored several representations for turning a raw event stream into something a neural network can process.
Dense approaches accumulate events into voxel grids or frame-like tensors and apply standard CNNs~\cite{maqueda2018event,zhu2019unsupervised,gehrig2019est}, sacrificing the sparsity and asynchrony that motivate event cameras.
Spiking neural networks (SNNs) instead process events one at a time, at the cost of training difficulty and limited scalability for deep architectures~\cite{lee2016training}.

Hand-crafted descriptors such as HATS~\cite{sironi2018hats}, which summarize local event history with time surfaces, represents an important early approach to event-based object classification.
Closer to our work, graph methods treat events directly as spatio-temporal point clouds, connecting nearby events into a graph and applying \acp{GNN}~\cite{bi2020graph,schaefer2022aegnn,qi2017pointnet++}.
In our work, we use PointNet++~\cite{qi2017pointnet++} as our primary backbone because it preserves per-event irregularity without forcing events onto a dense grid.
More recently, state-space sequence models in the style of Mamba have been explored for event-based vision, including EventMamba~\cite{Ge2025aaai}.
We test whether our recipe transfers to this structurally different, non-graph-based backbone family.

Where the event rate appears in prior pipelines, it is treated as a nuisance variable to be normalized away rather than as a signal to be exploited~\cite{zhu2019motion,schaefer2022aegnn,bi2020graph}.
One common strategy rescales each sample's timestamps into a fixed temporal window before any learning happens~\cite{zhu2019motion,schaefer2022aegnn}.
For instance, AEGNN maps every recording onto a common 300\,ms window regardless of its original duration~\cite{schaefer2022aegnn}, which by construction removes the original event rate from the data before the model ever sees it.
A second strategy leaves timestamps untouched at test time but randomly rescales the time axis during training, so that the network is explicitly encouraged to produce the same prediction regardless of the event rate of the scene~\cite{bi2020graph,zhu2019eventgan}.
Both strategies share the same underlying assumption: that the event rate is a nuisance to be eliminated, either by preprocessing or by invariance training.
We argue and show empirically that this assumption is costly whenever the event rate is class-relevant, since both strategies discard label information that a rate-aware model could otherwise use.

Feature-wise Linear Modulation (FiLM)~\cite{perez2018film} predicts a per-channel affine transform, consisting of a scale and a shift, from an auxiliary conditioning input, and applies it to intermediate feature maps.

Curriculum learning introduces training examples or tasks in a staged manner, often progressing from easier to more difficult settings~\cite{bengio2009curriculum}.
Our work adopts this paradigm by first learning under natural event rates and subsequently introducing temporal rate augmentation.

Test-time adaptation (TTA) methods adjust a trained model using only unlabeled test input to close the gap between training and deployment distributions, \eg{}, TENT~\cite{wang2021tent}, which minimizes prediction entropy online.
In contrast, our work adopts an offline, parameter-efficient adaptation strategy tailored to event-rate shifts~\cite{iwasawa2021test}.

\section{Method}
\label{sec:method}

\begin{figure}[t]
    \centering
    \includegraphics[width=\textwidth]{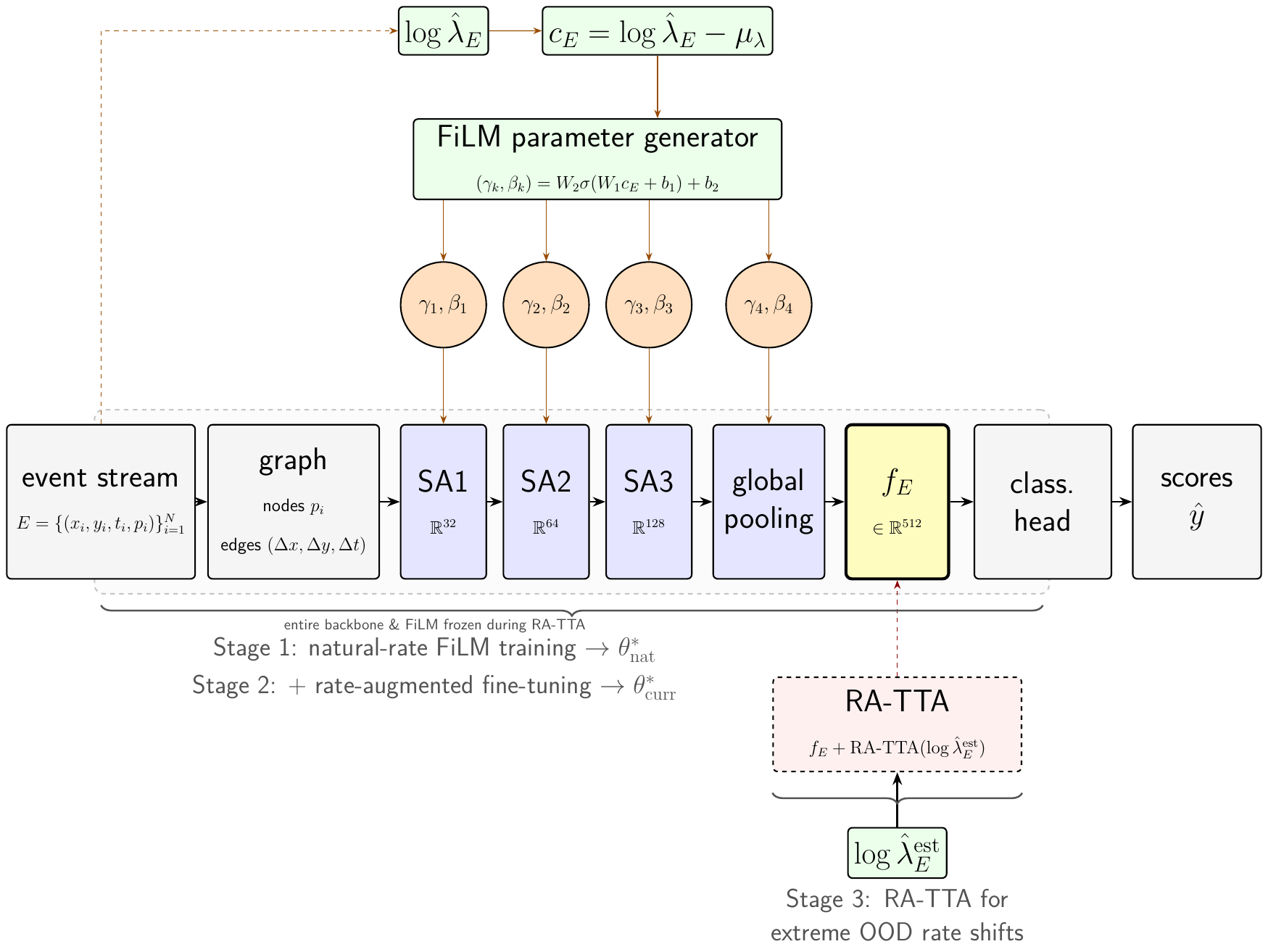}
    \caption{Overview of the proposed FiLM-Rate framework.
    An event stream is converted into a spatio-temporal radius graph and processed by a PointNet++~\cite{qi2017pointnet++} backbone.
    FiLM layers~\cite{perez2018film} condition the intermediate representations using a centered log-rate signal after each set-abstraction stage (SA) and after global pooling.
    The resulting pooled graph descriptor, $f_E$, is used for classification and can optionally be augmented at test time using RA-TTA (rate-adapter test-time adaptation) to improve robustness to extreme event-rate shifts while keeping the backbone frozen.}
    \label{fig:architecture}
\end{figure}

We describe the method in four parts: the graph representation and backbone we build on (\cref{sec:method:backbone});
how we condition that backbone on the event rate (\cref{sec:method:film}); the two-stage training procedure that combines rate conditioning with robustness to rate shifts (\cref{sec:method:curriculum});
and a small test-time module for shifts in the event rate, larger than any seen in training (\cref{sec:method:rptta}).
An overview of our proposed architecture is shown in~\cref{fig:architecture}.

\subsection{Graph representation and backbone}
\label{sec:method:backbone}

This section describes how a raw event stream is turned into a graph, the network that uses the graph for classification, and the scalar signal we will later use to condition the network on the recording's event rate.

A sample is a set of $N$ events recorded within a fixed time window, $E = \{(x_i, y_i, t_i, p_i)\}_{i=1}^{N}$, where $(x_i, y_i)$ is the pixel location, $p_i$ the polarity (sign of the brightness change), and $t_i$ the timestamp of the $i$-th event.
We build a spatio-temporal radius graph over these events:
each event is a node, its feature is its polarity $p_i$, and an edge connects two events that lie within a fixed spatial and temporal radius of each other.
Each edge carries the normalized spatial and temporal offsets between the two events it connects, $(\Delta x/r_{xy},\, \Delta y/r_{xy},\, \Delta t/r_t)$, where $\Delta x$ and $\Delta y$ are the distances between the two events in $x$ and $y$, and $\Delta t$ is their timestamp difference.
$r_{xy}$ and $r_t$ are the spatial and temporal radii used to build the graph.

We use this graph to classify the input to one of $C$ object categories ($C=101$ for N-Caltech101) using a PointNet++~\cite{qi2017pointnet++} architecture.
The network has three \emph{set-abstraction} (SA) stages, each of which groups nearby events into local neighbourhoods and summarizes each neighbourhood into a feature vector, at feature dimensionalities $32$, $64$, and $128$ for the first, second, and third stage respectively.
A \emph{global pooling layer} then summarizes the whole graph into a single descriptor $f_E \in \mathbb{R}^{512}$, by concatenating the elementwise maximum and mean over all remaining node features.
A final classification head, a small fully connected network with a hidden layer of size 128, maps this descriptor to class scores.
The whole architecture is visualized in~\cref{fig:architecture}.
In total, this backbone has 163{,}429 parameters which we refer to as the \emph{baseline budget}.

To later condition this network on the recording's event rate, we define a per-recording log-rate,
\begin{equation}
\loglam_E = \log\frac{N}{\max_i(t_i) - \min_i(t_i)},
\label{eq:rate}
\end{equation}
\ie{}, the logarithm of the number of events in the recording divided by its duration.
We compute exactly one event-rate value for each input recording, equivalent to one value per sample in the dataset.
We center this quantity before using it as an input,
\begin{equation}
c_E = \loglam_E - \mu_\lambda,
\label{eq:rate-centered}
\end{equation}
where $\mu_\lambda$ denotes the mean of $\loglam_E$ computed over the
training set.
$c_E$ is the conditioning signal used throughout the rest of this paper.

\subsection{FiLM-Rate: per-layer conditioning}
\label{sec:method:film}

We make the backbone described above sensitive to the event rate by inserting a small conditioning layer after each set-abstraction stage and after the global pooling layer, shown in~\cref{fig:architecture}, each driven by the centered log-rate $c_E$ defined in~\cref{eq:rate-centered}.
Concretely, we use Feature-wise Linear Modulation (FiLM)~\cite{perez2018film}.
Given a feature vector $h \in \mathbb{R}^d$ produced at one of these points in the network and the conditioning signal $c$, a FiLM layer computes
\begin{equation}
\mathrm{FiLM}(h, c) = \bigl(1 + \gamma(c)\bigr) \odot h + \beta(c),
\qquad
(\gamma, \beta) = W_2\,\sigma(W_1 c + b_1) + b_2,
\label{eq:film}
\end{equation}
where $\gamma$ and $\beta$ are a per-channel scale and shift predicted from $c$ by a two-layer network with weights $W_1 \in \mathbb{R}^{32\times1}$, $W_2 \in \mathbb{R}^{2d\times32}$, biases $b_1 \in \mathbb{R}^{32}$, $b_2 \in \mathbb{R}^{2d}$, and a Gaussian Error Linear Unit (GELU) nonlinearity $\sigma$, where $\odot$ denotes elementwise multiplication.
We initialize $W_2$ and $b_2$ to zero, while leaving $W_1$ and $b_1$ at their default initialization, so that at the start of training $(\gamma,\beta)\equiv(0,0)$ and~\cref{eq:film} reduces to the identity map $h\mapsto h$, so the network only departs from it as the conditioning parameters are learned.

The four FiLM layers together add 48{,}832 parameters (212{,}261 total, ${\sim}30\%$ over the baseline budget).

\subsection{Curriculum: a two-stage training procedure}
\label{sec:method:curriculum}

The conditioned network from~\cref{sec:method:film} learns to use the event rates observed during training.
However, deployment may expose the model to event rates that lie outside the distribution of the training data.
To improve robustness to out-of-distribution (OOD) event rates, a common strategy is \emph{rate augmentation}:
during training, each recording's timestamps are rescaled by a randomly chosen factor $\rho$ before the graph is built, so that the network is repeatedly exposed to the same recording at several different apparent rates.
Concretely, for a rescaling factor $\rho$ we define the operator
\begin{equation}
T_\rho(E) = \{(x_i,\, y_i,\, t_i/\rho,\, p_i)\}_{i=1}^{N},
\qquad
\loglam_{T_\rho(E)} = \loglam_E + \log\rho,
\label{eq:rho}
\end{equation}
which rescales every timestamp in $E$ by $1/\rho$ and updates the log-rate of~\cref{eq:rate} accordingly, so that the conditioning signal given to the network stays consistent with the rescaled input.

We combine rate conditioning and rate augmentation in two successive training stages, which together we call our curriculum.
Writing $F_\theta$ for the network of~\cref{sec:method:backbone,sec:method:film} with parameters $\theta$, and $\mathcal{L}_\mathrm{CE}$ for the cross-entropy loss, \emph{natural-rate training} first optimizes on natural, unrescaled recordings only, giving the natural-rate solution $\theta^*_\mathrm{nat}$:

\begin{equation}
\theta^*_\mathrm{nat} = \arg\min_\theta\;
\mathbb{E}_{(E,y)\sim\mathcal{D}}\;
\mathcal{L}_\mathrm{CE}\bigl(F_\theta(E),\, y\bigr).
\label{eq:stage1}
\end{equation}

\emph{Rate-augmented fine-tuning} then continues training the same network
from $\theta^*_\mathrm{nat}$, applying the rescaling of~\cref{eq:rho} at every
step with $\rho$ drawn uniformly from
$\mathcal{R} = \{0.5, 1.0, 2.0\}$:
\begin{equation}
\theta^*_\mathrm{curr} = \arg\min_\theta\Bigr|_{\theta_0 = \theta^*_\mathrm{nat}}\;
\mathbb{E}_{(E,y)\sim\mathcal{D},\,\rho\sim\mathcal{R}}\;
\mathcal{L}_\mathrm{CE}\bigl(F_\theta(T_\rho(E)),\, y\bigr).
\label{eq:stage2}
\end{equation}
only the starting point differs.\footnote{Implementation note: the per-recording event rate metadata used to compute~\cref{eq:rate} must survive every reconstruction of the graph data object during augmentation and subsampling;
if it silently defaults to zero, the conditioned network trains on a constant signal and collapses.
Our implementation keeps this metadata through all such paths.}

We compare this two-stage curriculum against training directly with rate augmentation from a random initialization in~\cref{sec:exp:ablation}, and against the same recipe ported to two other backbone families in~\cref{sec:multi-arch}.

\subsection{RA-TTA: a rate adapter for extreme shifts}
\label{sec:method:rptta}

The curriculum of~\cref{sec:method:curriculum} improves robustness to the event-rate shifts encountered during training.
However, deployment can produce rate shifts well outside this range.
For such cases, we add a small module, trained separately after the main network and left inactive unless it is explicitly enabled.
This module adjusts the pooled descriptor $f_E$ based on the event rate observed at test time, shown in~\cref{fig:architecture}.
We refer to this module as a \emph{rate adapter}.
Formally,
\begin{equation}
\tilde f_E = f_E + \alpha \cdot \mathrm{MLP}_\mathrm{adapt}(\loglam_E^{\mathrm{est}}),
\label{eq:rptta}
\end{equation}
where $\loglam^{\mathrm{est}}_E$ is the log-rate of~\cref{eq:rate} computed from the events actually observed at test time, $\mathrm{MLP}_\mathrm{adapt}$ is a two-layer multilayer perceptron (MLP) mapping this scalar to a vector in $\mathbb{R}^{512}$ (about 18K parameters, under 10\% of the base network's size), and the scalar gate $\alpha$ is zero-initialized so that the adapter has no effect before it is trained.
We call this procedure RA-TTA (rate-adapter test-time adaptation).
The FiLM-conditioned base network is kept frozen throughout the training of the adapter.
Only the adapter network and $\alpha$ are updated, using temporal rescaling factors ranging from $0.125$ to $8.0$, so that the adapter is exposed to event-rate shifts far beyond those encountered during the training of the base network.

\section{Experiments}
\label{sec:experiments}

We begin by describing the datasets, pre-processing pipeline, training configuration, and evaluation protocols used throughout the paper (\cref{sec:exp:setup}).
We then place our results in the context of prior work on N-Caltech101 (\cref{sec:exp:context}), present the main quantitative results (\cref{sec:exp:main}), and analyze the contributions of the individual components through a series of controlled ablations (\cref{sec:exp:ablation}).
We then test two alternative explanations for the gain in performance: that it is a shortcut driven by a rate/class correlation (\cref{sec:exp:shortcut}), and that it is an artifact of one specific backbone (\cref{sec:multi-arch}).
We also evaluate across multiple datasets (\cref{sec:multi}).

\subsection{Setup}
\label{sec:exp:setup}

\paragraph{Data \& preprocessing.}
We use the dataset N-Caltech101~\cite{orchard2015converting} for our experiments.
It contains $101$ classes, and after pre-processing, we use $6{,}967$ samples for training and $1{,}742$ samples for test.
Our pipeline preserves the natural timestamps (no per-recording normalization) and stores the per-recording event rate metadata.
The test set spans $\loglam \in [10.79, 14.05]$, corresponding to an approximately $26\times$ variation in event rate across recordings.
Event streams are represented as spatio-temporal radius graphs with radius $r=6$ (a 1\,ms displacement is treated numerically as equivalent to a 1-pixel displacement during construction), a maximum node degree of $32$, and at most $10{,}000$ events per sample.

\paragraph{Training.}
All models are trained with Adam (initial learning rate $10^{-3}$, batch size 16) for up to $80$ epochs, with the learning rate reduced by $10\times$ after epoch $30$ and early stopping (patience $15$) to prevent overfitting.
The RA-TTA module is trained separately for 30 epochs after the base model has converged.
All experiments run on a single NVIDIA A4000 GPU (16\,GB), with a typical training run taking $3$--$5$ hours.

\paragraph{Evaluation protocols.}
We evaluate all methods under two complementary protocols.
First, a deployment-oriented \emph{natural-rate} setting sorts the test recordings by event rate ($\loglam$) into seven equi-quantile buckets (about 249 recordings per bucket for N-Caltech101), reporting mean, worst-case, and per-bucket accuracy across the full range of naturally occurring event rates.

Second, we evaluate robustness to synthetic event-rate shifts by rescaling timestamps using factors $\rho \in \{0.125, 0.25, 0.5, 1.0, 2.0, 4.0, 8.0\}$ and rebuilding the corresponding graphs, reporting the mean and worst-case accuracy across all rescaling factors and the native-rate performance at $\rho=1.0$.

\subsection{Results on N-Caltech101}
\label{sec:exp:context}
\begin{table}[t]
\centering
\caption{Top-1 accuracy on N-Caltech101.
This is not a controlled comparison: protocols differ in splits, pre-processing, input representation, and pre-training.
Non-graph methods are shown in gray.
Numbers from the original papers are taken from~\cite{wang2022hardvs};
Parameter counts for RG-CNN, AEGNN, EV-VGCNN, and VMV-GCN are as measured on N-Caltech101 in~\cite{xie2024evstr}.
Our models use no pre-training and a deliberately small backbone.
}
\label{tab:context}
\small
\setlength{\tabcolsep}{5pt}
\begin{tabular}{llcc}
\toprule
Method & Representation & Params $\downarrow$ & Top-1 (\%) $\uparrow$ \\
\midrule
\textcolor{gray}{EventNet~\cite{sekikawa2019eventnet}} & \textcolor{gray}{point} & \textcolor{gray}{--} & \textcolor{gray}{42.5} \\
\textcolor{gray}{HATS~\cite{sironi2018hats}} & \textcolor{gray}{time surfaces} & \textcolor{gray}{--} & \textcolor{gray}{64.2} \\
\textcolor{gray}{EST~\cite{gehrig2019est}} & \textcolor{gray}{learned grid + CNN} & \textcolor{gray}{${\sim}$21M} & \textcolor{gray}{75.3$^\dagger$} \\
RG-CNN~\cite{bi2020graph} & graph & 19.46M & 65.7 \\
AEGNN~\cite{schaefer2022aegnn} & graph & 20.40M & 66.8 \\
EV-VGCNN~\cite{deng2022evvgcnn} & voxel graph & 0.84M & 74.8 \\
VMV-GCN~\cite{xie2022vmvgcn} & voxel graph & 0.86M & 77.8 \\
\midrule
PointNet++ baseline & graph & 0.16M & 60.9 \\
\textbf{FiLM-Rate curriculum (ours)} & graph & 0.21M & \textbf{67.2} \\
\bottomrule
\multicolumn{4}{l}{\footnotesize $^\dagger$Higher figures have been reported with ImageNet pre-training; see~\cite{gehrig2019est}.}
\end{tabular}
\end{table}
\Cref{tab:context} shows two main findings.
First, dense or voxel-graph methods with large CNN backbones and pre-training (EST, EV-VGCNN, VMV-GCN) exceed our absolute accuracy.
However, we do not make any state-of-the-art claims.
Second, within the lightweight-graph regime, our recipe lifts a 0.16M-parameter backbone from 60.9\% to 67.2\%, to AEGNN-level accuracy at a small fraction of the parameters \emph{without touching the architecture's capacity for event processing}.
All subsequent comparisons in this section are internally controlled: identical backbone, data, and schedule.

\subsection{Main results}
\label{sec:exp:main}

\begin{table}[t]
\centering
\caption{Main results on N-Caltech101 (identical backbone, data, and schedule).
We report Top-1 accuracy, aggregated as the average and worst-case across conditions~\cite{hendrycks2019benchmarking,sagawa2020distributionally}.
``Natural-rate'' aggregates over 7 equi-quantile natural-rate buckets.
``Synthetic-rate'' aggregates over 7 synthetic factors $\{0.125,\dots,8.0\}$; $1.0\times$ = native rate.
\emph{R-Canon} denotes a baseline that learns a per-recording rescaling factor to canonicalize the event rate before a frozen classifier.
\best{Bold} best, \second{underline} second best.}
%
\label{tab:main}
\small
\setlength{\tabcolsep}{2pt}
\begin{tabular}{lcccccc}
\toprule
& \multicolumn{2}{c}{Natural-rate $\uparrow$} & \multicolumn{2}{c}{Synthetic-rate $\uparrow$} & & \\
\cmidrule(lr){2-3} \cmidrule(lr){4-5}
Method & Avg. & Worst & Avg. & Worst & 1.0$\times \uparrow$ & Params $\downarrow$\\
\midrule
Baseline                   & 60.9 & 50.0 & 16.0 & 3.4  & 60.6 & 163K \\
Baseline widehead          & 62.0 & 50.8 & 16.0 & 2.4  & 61.9 & 208K \\
RateConditioned (head)     & 60.0 & 52.0 & 17.5 & 5.3  & 59.9 & 163K \\
ra-aug-only                & 60.3 & 49.2 & 47.0 & 8.0  & 59.8 & 163K \\
R-Canon on baseline        & 44.7 & 34.3 & 25.2 & 8.6  & 44.8 & 163K+264K \\
RA-TTA on baseline         & 40.2 & 29.4 & 30.5 & 13.9 & 40.1 & 163K+18K \\
RA-TTA on ra-aug-only      & 59.0 & 46.0 & 52.4 & 38.1 & 58.9 & 163K+18K \\
\midrule
FiLM-Global-Only (no aug)  & 60.3 & 51.2 & 16.4 & 5.0  & 60.7 & 197K \\
FiLM-Rate (no aug)         & \second{64.4} & \best{55.7} & 20.4 & 5.2 & 63.4 & 212K \\
FiLM-Rate (ra-aug)         & 61.7 & 51.6 & 47.4 & 10.5 & 61.3 & 212K \\
\textbf{FiLM-Rate (curriculum)} & \best{67.2} & 53.6 & 50.3 & 12.7 & \best{67.6} & 212K \\
RA-TTA on FiLM (ra-aug)    & 60.3 & 48.8 & \second{54.3} & \best{44.7} & 60.3 & 212K+18K \\
\textbf{RA-TTA on FiLM (curr.)} & 64.5 & 50.8 & \best{57.5} & \second{41.7} & \second{64.8} & 212K+18K \\
\bottomrule
\end{tabular}
\end{table}

\begin{figure}[!t]
\centering
\includegraphics[width=\textwidth]{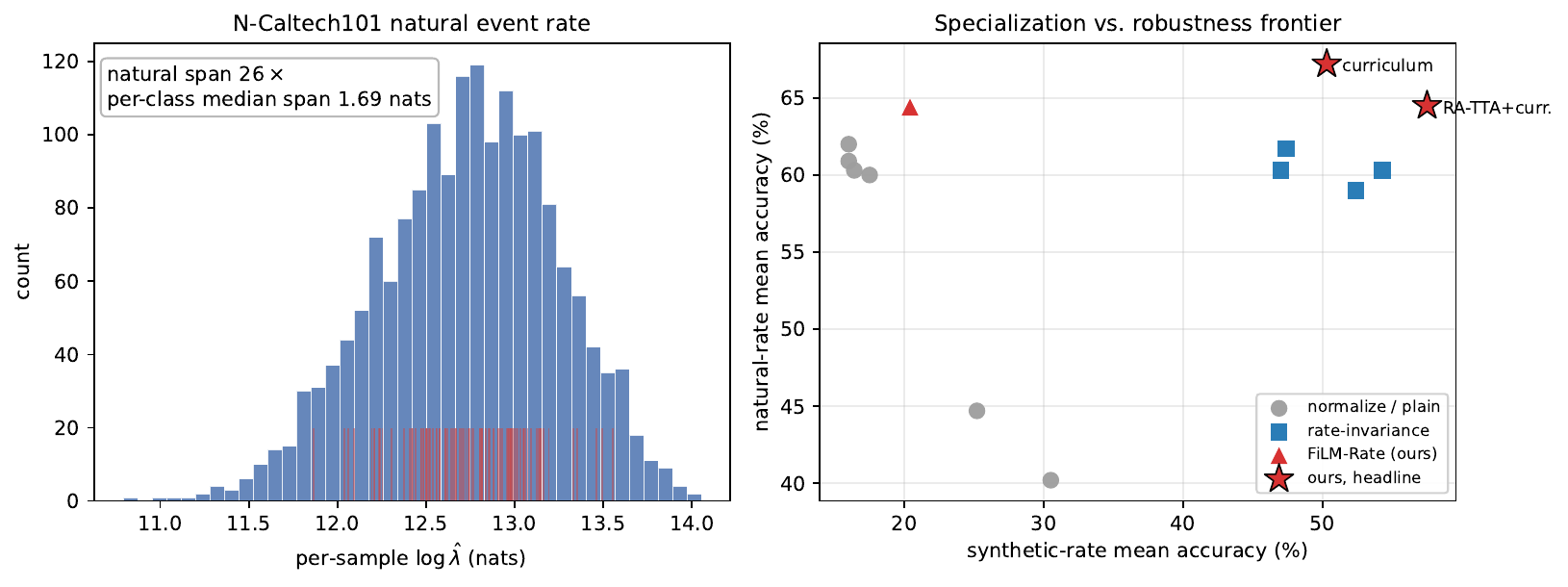}
\caption{\textbf{Left:} Distribution of recording-level event rates on N-Caltech101.
Red vertical markers indicate the median event rate of each of the 101 classes (marker height has no meaning and is shown uniformly for clarity).
The figure illustrates that the event rate varies substantially across object categories.
The natural spread is ${\sim}26\times$ and depends strongly on the class, yet standard normalization collapses it to a constant.
\textbf{Right:} natural-rate vs.\ synthetic-rate mean accuracy of every method in~\cref{tab:main}.
Rate-invariance training moves right but down; our curriculum and its RA-TTA extension sit on the upper frontier.}
\label{fig:pareto}
\end{figure}

The main results are listed in~\cref{tab:main}.
On the \textit{natural-rate} evaluation, our method, FiLM-Rate Curriculum, reaches \textbf{67.2\%} mean accuracy against 60.9\% for the baseline, a gain of \textbf{6.3\,pp} on an identical backbone.
The gain is distributed across the seven event-rate quantiles defined in~\cref{sec:exp:setup}, rather than arising from a single subset of the test set (see~\cref{tab:bucket} in the supplementary material).
The curriculum improves over FiLM (no aug) in $6$ out of $7$ buckets, with per-bucket gains of $2.4$--$10.5\,pp$.
The only exception is the highest rate bucket (b6), where the curriculum trades 2.1\,pp for improved robustness to event-rate shifts.
Notably, our method achieves its strongest performance on the unmodified test recordings, indicating that the gains observed under synthetic event-rate shifts do not come at the expense of natural-rate performance.
\Cref{fig:pareto} (right) shows that RA-TTA on FiLM (rate-augmented) and RA-TTA on FiLM (curriculum) define the natural-versus-synthetic Pareto frontier.

\subsection{Ablations: where does the gain come from?}
\label{sec:exp:ablation}

\begin{table}[t]
\centering
\caption{Consolidated ablations. \textbf{Top:} capacity vs.\ mechanism vs.\ placement: matched extra parameters achieve +1.1\,pp, per-layer rate conditioning additional +2.4\,pp, while conditioning only the pooled feature representation $f_E$ (\ie{}, after global pooling) provides no benefit.
\textbf{Bottom:} the curriculum's gain comes from the warm-start initialization, not the augmentation range.
Bold marks the best value in each block.}
\label{tab:ablations}
\small
\setlength{\tabcolsep}{5pt}
\begin{tabular}{llccc}
\toprule
\multicolumn{2}{l}{\emph{Capacity / placement}} & $\Delta$ params $\downarrow$ & Nat.\ mean $\uparrow$ & $\Delta \uparrow$ \\
\midrule
\multicolumn{2}{l}{Baseline}            & --     & 60.9 & --     \\
\multicolumn{2}{l}{Baseline widehead}   & +44K   & 62.0 & +1.1   \\
\multicolumn{2}{l}{FiLM-Global-Only}    & +34K   & 60.3 & $-$0.6 \\
\multicolumn{2}{l}{\textbf{FiLM-Rate (per-layer)}} & +49K & \best{64.4} & \best{+3.5} \\
\midrule
\emph{Curriculum source} & Init & Aug range & Nat.\ mean $\uparrow$ & 1.0$\times \uparrow$ \\
\midrule
FiLM (no aug)       & scratch       & $\{1.0\}$           & 64.4 & 63.4 \\
FiLM (mild aug)     & scratch       & $\{0.7, 1.0, 1.5\}$ & 61.3 & 62.0 \\
FiLM (default aug)  & scratch       & $\{0.5, 1.0, 2.0\}$ & 61.7 & 61.3 \\
\textbf{FiLM (curriculum)} & FiLM (no aug) & $\{0.5, 1.0, 2.0\}$ & \best{67.2} & \best{67.6} \\
\bottomrule
\end{tabular}
\end{table}

\paragraph{Capacity is not the reason.}
A width-matched baseline (\emph{Baseline widehead}) gains only about a third of FiLM's improvement (\cref{tab:ablations}, top), showing that most of the gain comes from rate conditioning rather than added capacity.

\paragraph{Placement of the modulation is important.}
FiLM applied only at the pooled feature representation $f_E$ (\ie{}, after global pooling) provides no measurable benefit despite containing most of the added parameters (\cref{tab:ablations}, top), suggesting that rate conditioning is most effective before local spatio-temporal information has been aggregated.

\paragraph{Initialization matters more than the augmentation range.}
A natural alternative hypothesis, that the curriculum merely finds a better augmentation range, is not supported by~\Cref{tab:ablations} (bottom): the curriculum outperforms every from-scratch variant by $\ge$5.5\,pp while using the identical default range.

\subsection{Could the gain just be a rate/class shortcut?}
\label{sec:exp:shortcut}

If N-Caltech101's event rate were strongly tied to class identity, a model that exploits rate could simply be exploiting a labeling shortcut, rather than learning genuine rate-conditional features.
We test this directly using the pre-subsampling per-recording log-rate ($\loglam_E$ of~\cref{eq:rate}, before the 10K-event cap) paired with class labels on the N-Caltech101 test split.
Rate does differ across classes: a standard test for whether group means differ (one-way ANOVA) is highly significant across the 101 classes ($F=9.78$, $p=1.2\times10^{-107}$), confirming rate is not class-independent, as we mentioned previously.
However, the size of this effect is what matters for the shortcut argument, and it is small: rate and class share only $6.6\%$ of the information needed to predict the class label (mutual information $I(\loglam_E; y) = 0.278$ against a class uncertainty of $H(y)=4.181$, both measured in nats), and the per-class average log-rate spans just $1.43$ (std.\ $0.302$).
Classifiers that see \emph{only} $\loglam_E$ make this concrete: against a most-frequent-class baseline of $9.18\%$, a rate-only logistic regression and MLP reach $16.48\%$ and $16.36\%$ Top-1 accuracy, respectively.
Far too low to explain our $6.3$\,pp gain over a $60.9\%$ baseline that already sees the full event stream.

\paragraph{Leakage and gain are anti-correlated.}
This weak rate/class coupling bounds the shortcut explanation on N-Caltech101 but does not rule it out in general, so we repeat the analysis on DVS128 Gesture, where our recipe gains the most (\cref{tab:multi}; full comparison in~\cref{tab:shortcut-supp} in the supplementary material).
There, event rate carries almost no class information: it accounts for only $1.75\%$ of the class uncertainty (vs.\ $6.6\%$ on N-Caltech101).
The per-class average log-rate spans just $0.11$ (vs.\ $1.43$), and a rate-only classifier scores $21.79\%$, exactly the value of the most-frequent class baseline (see~\cref{tab:shortcut-supp} in the supplementary material).
Thus, it learns nothing beyond guessing the most common class.
Yet this is exactly the dataset where the gain is largest ($+14.0$\,pp, more than double N-Caltech101's).
If the gain were a rate/class shortcut, it should grow with how much rate leaks about the label.
Instead, the relationship is inverted, which rules out that kind of leakage as the explanation.
This agrees with the placement ablation of~\cref{sec:exp:ablation}: conditioning after pooling, where a leaked label would be easiest to read off, gives no gain.
While conditioning at the neighborhood-forming stages before, where it can change how events are grouped, yields the largest effect.

\section{What Is FiLM Actually Computing?}
\label{sec:feature-analysis}

We now verify that FiLM uses the rate signal in the way we intend. You can find a corresponding visualization in~\cref{fig:film} in the supplementary material.
Extracting the learned $(\gamma_k, \beta_k)$ per FiLM layer and grouping them by the natural-rate buckets (\cref{fig:film}, left) shows both multiplicative and additive modulation are active at \emph{every} layer and vary with rate; different rates trigger different processing, exactly the capability the variant which only modulates the pooled feature representation $f_E$ (\ie{}, after global pooling) lacks.
In the pooled feature representation $f_E$, rate buckets do \emph{not} form isolated clusters.
FiLM is modulatory, reshaping feature scales conditional on the event rate, rather than injecting rate as a competing axis that displaces class structure.
Quantitatively (\cref{fig:film}, right), the curriculum's per-bucket class silhouette dominates the baseline's in all seven buckets, indicating the head receives more discriminative information even where the argmax does not change.


\section{Cross-Dataset Evaluation}
\label{sec:multi}

\begin{table}[tbh]
\centering
\caption{Natural-rate mean accuracy across four benchmarks with identical architecture, schedule, and pre-processing schema; only per-sensor set-abstraction (SA) radii differ.
	Bold marks the best value per column.}
\label{tab:multi}
\footnotesize
\setlength{\tabcolsep}{2pt}
\begin{tabular}{lcccc}
\toprule
Method & DVS128 Gesture & N-Caltech101 & CIFAR10-DVS & N-Cars \\
       & (11-cls) & (101-cls) & (10-cls) & (2-cls, sat.) \\
\midrule
Baseline               & 68.6 & 60.9 & 51.5 & \best{93.9} \\
FiLM-Rate (no aug)     & 71.2 & 64.4 & 53.2 & 93.6 \\
\textbf{FiLM-Rate (curr.)} & \best{82.5} & \best{67.2} & \best{56.7} & 93.5 \\
\midrule
$\Delta$ vs.\ baseline & +14.0\,pp & +6.3\,pp & +5.2\,pp & $-$0.4\,pp$^*$ \\
\bottomrule
\addlinespace[2pt]
\multicolumn{5}{l}{\scriptsize $^*$Robustness improves instead: 73.5\% vs.\ 61.7\% at $0.125\times$; synth.\ mean 87.9\% vs.\ 84.9\%.}
\end{tabular}
\end{table}

\paragraph{DVS128 Gesture~\cite{amir2017lowpower}.}
This dataset uses a different sensor geometry ($128\times128$), task family (11-class gesture), and a substantially different event-rate distribution.
The curriculum gains are \textbf{+14.0\,pp} over the baseline, listed in~\cref{tab:multi}, larger than on N-Caltech101 with the same ordering (baseline ${<}$ FiLM-no-aug ${<}$ curriculum).
This shows that our proposed architecture is not specific to the N-Caltech101 dataset.
\Cref{sec:multi-arch:lighting} further stratifies this dataset's evaluation by its five native recording-lighting conditions, addressing the concern that gains reported under one lighting setup may not reflect deployment-realistic sensor conditions.

\paragraph{CIFAR10-DVS~\cite{li2017cifar10dvs}.}
A third sensor and recording protocol (10-class, converted from CIFAR-10 by repeated saccadic motion). The curriculum gains $+5.2$\,pp over the baseline ($51.5 \rightarrow 56.7$) with worst-bucket accuracy improving from $44.1\%$ to $51.1\%$, again with the same ordering as on DVS128 Gesture: FiLM-no-aug's accuracy falls strictly between the baseline and the curriculum.

\paragraph{N-Cars~\cite{sironi2018hats}.}
N-Cars contains only two classes.
Our backbone reaches approximately 94\% on natural recordings (\cref{tab:multi}), and all methods perform similarly.
This is the one case in our study where rate conditioning does not help.
This is predicted rather than anomalous.
N-Cars has by far the widest event-rate spread we study ($4.77$ nats, ${\approx}118\times$).
Thus, a method that benefited from rate variation as such should benefit most here.
It does not: almost none of N-Cars' spread is class-informative, so rate conditioning has nothing to exploit.
Together with the leakage/gain comparison above, this gives a consistent picture: the benefit of rate conditioning tracks the fraction of rate variance that separates classes, not the size of the spread (full variance decomposition across all four datasets in~\cref{tab:variance-supp} of the supplementary material).
Robustness to synthetic shifts still improves (worst-case synthetic accuracy $61.7\% \rightarrow 73.5\%$): the conditioning gain is gated by class-informative rate variance, while the augmentation-driven robustness gain transfers regardless.


\section{Cross-Architecture Generalization}
\label{sec:multi-arch}

The results above show that FiLM-Rate conditioning with the two-stage curriculum improves one specific backbone, PointNet++.
To test generality, we port the identical recipe, unchanged, onto EventMamba~\cite{Ge2025aaai}, a state-space point-sequence model structurally unlike PointNet++ (no spatial graph at all), evaluated on DVS128 Gesture.

\begin{table}[t]
\centering
\caption{Cross-architecture summary: does the FiLM-Rate curriculum improve natural-rate Top-1 accuracy over an unconditioned baseline of the same architecture? PointNet++ is evaluated on N-Caltech101; EventMamba on DVS128 Gesture.
	Bold marks the better (FiLM-Rate curriculum) value per row.}
\label{tab:crossarch}
\small
\setlength{\tabcolsep}{5pt}
\begin{tabular}{lccc}
\toprule
Backbone & Baseline & FiLM-Rate (curr.) & $\Delta$ \\
\midrule
PointNet++ (\cref{sec:exp:main})   & 60.9  & \best{67.2}  & $+6.3$\,pp \\
EventMamba (state-space)           & 67.4  & \best{74.1}  & $+6.7$\,pp \\
\bottomrule
\addlinespace[2pt]

\end{tabular}
\end{table}

\paragraph{EventMamba (state-space).}
We apply the same recipe to a Mamba-based point-sequence classifier on DVS128 Gesture, adding FiLM after each state-space stage.
The recipe transfers here too: mean accuracy follows the same ordering as~\cref{tab:main} (baseline $67.39\% <$ FiLM-Rate no-aug $70.24\% <$ curriculum $\mathbf{74.11\%}$).
With the same ordering for the worst rate bucket, a gain of $+6.7$\,pp mean and $+8.9$\,pp worst-bucket, summarized alongside the PointNet++ result in~\cref{tab:crossarch}.

\label{sec:multi-arch:lighting}
\paragraph{Lighting-stratified robustness (DVS128 Gesture).}

DVS128 Gesture tags each recording with its lighting condition.
Illumination can affect when events are triggered, making it a relevant nuisance factor.
We stratify the EventMamba results by lighting condition.
The curriculum improves over the baseline under every well-populated condition, with the largest gain in the lab setting ($+9.60$\,pp).
The full breakdown is reported in~\cref{tab:lighting-supp} in the supplementary material.
A similar pattern is observed for N-Cars~\cref{sec:multi}.

\section{Limitations, Future Work, and Conclusion}
\label{sec:limits}

\paragraph{Limitations and future work.}
Despite these encouraging results, there is further space for improvement.
First, the curriculum incurs a small loss at the highest observed event rates; stronger rate augmentation in the second stage might address this.
Second, RA-TTA trades a modest reduction in worst-case natural-rate performance for robustness to extreme event-rate shifts.
Third, we use the exact event rate computed from the complete recording; deployment on continuous streams would need an online estimate (\eg{}, a sliding window), which we expect FiLM-Rate to extend to naturally, though sensitivity to noisy estimates remains for future work.

\paragraph{Conclusion.}
The variation in the event rate is a fundamental characteristic of event streams from event cameras, rather than a nuisance to be removed.
Across multiple datasets, we show that the event rate carries class-relevant information that is often discarded by standard preprocessing pipelines.
By conditioning a graph-based backbone on this signal with a two-stage training curriculum, we obtain consistent gains in natural-rate accuracy (+6.3\,pp on N-Caltech101, +14.0\,pp on DVS128 Gesture) and in robustness to synthetic event-rate shifts, without sacrificing natural-rate performance, at a cost of under $30\%$ additional parameters.
Two contributions should outlast this specific work: the \emph{natural-rate evaluation protocol}, which exposes brittleness that standard synthetic-shift benchmarks hide, and the \emph{curriculum-as-initialization} recipe, which obtains specialization and robustness without the usual trade-off between them.

\bibliographystyle{splncs04}
\bibliography{main}

@article{lichtsteiner2008,
  author  = {Lichtsteiner, Patrick and Posch, Christoph and Delbruck, Tobi},
  title   = {A 128$\times$128 120 dB 15 $\mu$s Latency Asynchronous Temporal Contrast Vision Sensor},
  journal = {IEEE Journal of Solid-State Circuits},
  volume  = {43},
  number  = {2},
  pages   = {566--576},
  year    = {2008},
  doi     = {10.1109/JSSC.2007.914337},
}

@inproceedings{schaefer2022aegnn,
  author    = {Schaefer, Simon and Gehrig, Daniel and Scaramuzza, Davide},
  title     = {{AEGNN}: Asynchronous Event-based Graph Neural Networks},
  booktitle = {Proceedings of the IEEE/CVF Conference on Computer Vision and Pattern Recognition (CVPR)},
  year      = {2022},
}

@article{bi2020graph,
  author  = {Bi, Yin and Chadha, Aaron and Abbas, Alhabib and Bourtsoulatze, Eirina and Andreopoulos, Yiannis},
  title   = {Graph-based Spatio-Temporal Feature Learning for Neuromorphic Vision Sensing},
  journal = {IEEE Transactions on Image Processing},
  volume  = {29},
  pages   = {9084--9098},
  year    = {2020},
}

@inproceedings{qi2017pointnet++,
  author    = {Qi, Charles Ruizhongtai and Yi, Li and Su, Hao and Guibas, Leonidas J.},
  title     = {{PointNet++}: Deep Hierarchical Feature Learning on Point Sets in a Metric Space},
  booktitle = {Advances in Neural Information Processing Systems (NeurIPS)},
  year      = {2017},
}

@article{orchard2015converting,
  author  = {Orchard, Garrick and Jayawant, Ajinkya and Cohen, Gregory K. and Thakor, Nitish},
  title   = {Converting Static Image Datasets to Spiking Neuromorphic Datasets Using Saccades},
  journal = {Frontiers in Neuroscience},
  volume  = {9},
  pages   = {437},
  year    = {2015},
  doi     = {10.3389/fnins.2015.00437},
}

@inproceedings{perez2018film,
  author    = {Perez, Ethan and Strub, Florian and de Vries, Harm and Dumoulin, Vincent and Courville, Aaron},
  title     = {{FiLM}: Visual Reasoning with a General Conditioning Layer},
  booktitle = {Proceedings of the AAAI Conference on Artificial Intelligence},
  number    = {1},
  year      = {2018},
  doi       = {10.1609/aaai.v32i1.11671},
}

@inproceedings{bengio2009curriculum,
  author    = {Bengio, Yoshua and Louradour, J\'er\^ome and Collobert, Ronan and Weston, Jason},
  title     = {Curriculum Learning},
  booktitle = {Proceedings of the 26th International Conference on Machine Learning (ICML)},
  pages     = {41--48},
  year      = {2009},
}

@inproceedings{wang2021tent,
  author    = {Wang, Dequan and Shelhamer, Evan and Liu, Shaoteng and Olshausen, Bruno and Darrell, Trevor},
  title     = {Tent: Fully Test-Time Adaptation by Entropy Minimization},
  booktitle = {International Conference on Learning Representations (ICLR)},
  year      = {2021},
}

@inproceedings{iwasawa2021test,
  author    = {Iwasawa, Yusuke and Matsuo, Yutaka},
  title     = {Test-Time Classifier Adjustment Module for Model-Agnostic Domain Generalization},
  booktitle = {Advances in Neural Information Processing Systems (NeurIPS)},
  year      = {2021},
}

@inproceedings{zhu2019unsupervised,
  author    = {Zhu, Alex Zihao and Yuan, Liangzhe and Chaney, Kenneth and Daniilidis, Kostas},
  title     = {Unsupervised Event-based Learning of Optical Flow, Depth, and Egomotion},
  booktitle = {Proceedings of the IEEE/CVF Conference on Computer Vision and Pattern Recognition (CVPR)},
  year      = {2019},
}

@inproceedings{maqueda2018event,
  author    = {Maqueda, Ana I. and Loquercio, Antonio and Gallego, Guillermo and Garc\'ia, Narciso and Scaramuzza, Davide},
  title     = {Event-based Vision Meets Deep Learning on Steering Prediction for Self-Driving Cars},
  booktitle = {Proceedings of the IEEE/CVF Conference on Computer Vision and Pattern Recognition (CVPR)},
  year      = {2018},
}

@article{lee2016training,
  author  = {Lee, Jun Haeng and Delbruck, Tobi and Pfeiffer, Michael},
  title   = {Training Deep Spiking Neural Networks Using Backpropagation},
  journal = {Frontiers in Neuroscience},
  volume  = {10},
  pages   = {508},
  year    = {2016},
  doi     = {10.3389/fnins.2016.00508},
}

@inproceedings{zhu2019eventgan,
  author    = {Zhu, Alex Zihao and Wang, Ziyun and Khant, Kaung and Daniilidis, Kostas},
  title     = {{EventGAN}: Leveraging Large Scale Image Datasets for Event Cameras},
  booktitle = {IEEE International Conference on Computational Photography (ICCP)},
  year      = {2021},
  note      = {arXiv:1912.01584},
}

@inproceedings{sironi2018hats,
  author    = {Sironi, Amos and Brambilla, Manuele and Bourdis, Nicolas and Lagorce, Xavier and Benosman, Ryad},
  title     = {{HATS}: Histograms of Averaged Time Surfaces for Robust Event-based Object Classification},
  booktitle = {Proceedings of the IEEE/CVF Conference on Computer Vision and Pattern Recognition (CVPR)},
  year      = {2018},
}

@inproceedings{amir2017lowpower,
  author    = {Amir, Arnon and Taba, Brian and Berg, David and Melano, Timothy and McKinstry, Jeffrey and Di Nolfo, Carmelo and Nayak, Tapan and Andreopoulos, Alexander and Garreau, Guillaume and Mendoza, Marcela and Kusnitz, Jeff and Debole, Michael and Esser, Steve and Delbruck, Tobi and Flickner, Myron and Modha, Dharmendra},
  title     = {A Low Power, Fully Event-Based Gesture Recognition System},
  booktitle = {Proceedings of the IEEE/CVF Conference on Computer Vision and Pattern Recognition (CVPR)},
  year      = {2017},
}

@article{li2017cifar10dvs,
  author  = {Li, Hongmin and Liu, Hanchao and Ji, Xiangyang and Li, Guoqi and Shi, Luping},
  title   = {{CIFAR10-DVS}: An Event-Stream Dataset for Object Classification},
  journal = {Frontiers in Neuroscience},
  volume  = {11},
  pages   = {309},
  year    = {2017},
  doi     = {10.3389/fnins.2017.00309},
}

@inproceedings{gehrig2019est,
  author    = {Gehrig, Daniel and Loquercio, Antonio and Derpanis, Konstantinos G. and Scaramuzza, Davide},
  title     = {End-to-End Learning of Representations for Asynchronous Event-Based Data},
  booktitle = {Proceedings of the IEEE/CVF International Conference on Computer Vision (ICCV)},
  pages     = {5632--5642},
  year      = {2019},
}

@inproceedings{sekikawa2019eventnet,
  author    = {Sekikawa, Yusuke and Hara, Kosuke and Saito, Hideo},
  title     = {{EventNet}: Asynchronous Recursive Event Processing},
  booktitle = {Proceedings of the IEEE/CVF Conference on Computer Vision and Pattern Recognition (CVPR)},
  year      = {2019},
}

@inproceedings{wang2022hardvs,
  author    = {Wang, Xiao and Wu, Zongzhen and Jiang, Bo and Bao, Zhimin and Zhu, Lin and Li, Guoqi and Wang, Yaowei and Tian, Yonghong},
  title     = {{HARDVS}: Revisiting Human Activity Recognition with Dynamic Vision Sensors},
  booktitle = {Proceedings of the AAAI Conference on Artificial Intelligence},
  volume    = {38},
  pages     = {5615--5623},
  year      = {2024},
}

@misc{zhu2019motion,
  author        = {Zhu, Alex Zihao and Wang, Ziyun and Daniilidis, Kostas},
  title         = {Motion Equivariant Networks for Event Cameras with the Temporal Normalization Transform},
  year          = {2019},
  eprint        = {1902.06820},
  archivePrefix = {arXiv},
  primaryClass  = {cs.CV},
}

@inproceedings{deng2022evvgcnn,
  author    = {Deng, Yongjian and Chen, Hao and Chen, Huiying and Li, Youfu},
  title     = {{EV-VGCNN}: A Voxel Graph CNN for Event-based Object Classification},
  booktitle = {Proceedings of the IEEE/CVF Conference on Computer Vision and Pattern Recognition (CVPR)},
  year      = {2022},
  note      = {arXiv:2106.00216},
}

@article{xie2022vmvgcn,
  author  = {Xie, Bochen and Deng, Yongjian and Shao, Zhanpeng and Liu, Hai and Li, Youfu},
  title   = {{VMV-GCN}: Volumetric Multi-View Based Graph {CNN} for Event Stream Classification},
  journal = {IEEE Robotics and Automation Letters},
  volume  = {7},
  number  = {2},
  pages   = {1976--1983},
  year    = {2022},
}

@inproceedings{hendrycks2019benchmarking,
  author    = {Hendrycks, Dan and Dietterich, Thomas},
  title     = {Benchmarking Neural Network Robustness to Common Corruptions and Perturbations},
  booktitle = {International Conference on Learning Representations (ICLR)},
  year      = {2019},
}

@inproceedings{sagawa2020distributionally,
  author    = {Sagawa, Shiori and Koh, Pang Wei and Hashimoto, Tatsunori B. and Liang, Percy},
  title     = {Distributionally Robust Neural Networks for Group Shifts: On the Importance of Regularization for Worst-Case Generalization},
  booktitle = {International Conference on Learning Representations (ICLR)},
  year      = {2020},
}

@article{xie2024evstr,
  author  = {Xie, Bochen and Deng, Yongjian and Shao, Zhanpeng and Xu, Qingsong and Li, Youfu},
  title   = {Event Voxel Set Transformer for Spatiotemporal Representation Learning on Event Streams},
  journal = {IEEE Transactions on Circuits and Systems for Video Technology},
  year    = {2024},
  note    = {arXiv:2303.03856},
}

@article{Ge2025aaai,
  author  = {Ge, Chengjie and Fu, Xueyang and He, Peng and Wang, Kunyu and Cao, Chengzhi and Zha, Zheng-Jun},
  title   = {{EventMamba}: Enhancing Spatio-Temporal Locality with State Space Models for Event-Based Video Reconstruction},
  journal = {Proceedings of the AAAI Conference on Artificial Intelligence},
  volume  = {39},
  number  = {3},
  pages   = {3104--3112},
  year    = {2025},
  doi     = {10.1609/aaai.v39i3.32319},
}

\clearpage
\setcounter{page}{1}

\title{Supplementary Material:\\
       Rate as a Feature, Not a Nuisance:\\
       Curriculum FiLM Training for Event-Based\\
       Graph Neural Networks}
\maketitle
\renewcommand{\thetable}{S\arabic{table}}
\renewcommand{\thefigure}{S\arabic{figure}}
\setcounter{section}{0}
\renewcommand{\thesection}{S\arabic{section}}
\renewcommand{\theHsection}{S.\arabic{section}}

This supplement expands five results from the main paper with full per-condition and per-dataset detail: rate/class leakage versus accuracy gain (\cref{sec:s1}), per-lighting-condition robustness (\cref{sec:s2}), per-bucket natural-rate accuracy (\cref{sec:s3}), FiLM parameter analysis (\cref{sec:s4}), and log-rate variance decomposition across datasets (\cref{sec:s5}).

\section{Rate/class leakage vs.\ accuracy gain, full comparison}\label{sec:s1}

\Cref{tab:shortcut-supp} expands the summary in~\cref{sec:exp:shortcut} (``Is the gain a rate/class confound or a rate-only shortcut? ''): the full side-by-side of N-Caltech101 and DVS128 Gesture, the two datasets where the leakage/gain anti-correlation argument is made.

\begin{table}[h]
\centering
\caption{Rate/class leakage versus accuracy gain, on the two datasets where the recipe helps.
	If the gain were a rate-label shortcut, it would grow with leakage.
	It does the opposite: on DVS128 Gesture a rate-only classifier is exactly at chance, and that is where the recipe gains most.}
\label{tab:shortcut-supp}
\small
\setlength{\tabcolsep}{5pt}
\begin{tabular}{lcc}
\toprule
 & N-Caltech101 & DVS128 Gesture \\
\midrule
$I(\loglam_E; y) / H(y)$              & 6.65\% & 1.75\% \\
Per-class mean log-rate spread        & 1.43 nats & 0.11 nats \\
\midrule
Chance (most-frequent class)          & 9.18 & 21.79 \\
Rate-only classifier (best)           & 16.48 & 21.79 \\
\quad gain over chance                & $+7.30$ & $\mathbf{+0.00}$ \\
\midrule
FiLM-Rate curriculum gain over baseline & $+6.3$\,pp & $\mathbf{+14.0}$\,\textbf{pp} \\
\bottomrule
\end{tabular}
\end{table}

\section{Per-lighting-condition robustness, full breakdown}\label{sec:s2}

\Cref{tab:lighting-supp} expands the summary in~\cref{sec:multi-arch} (cross-architecture generalization): per-lighting-condition Top-1 accuracy on DVS128 Gesture using the EventMamba backbone, stratified by the recording's native illumination tag.

\begin{table}[h]
\centering
\caption{Per-lighting-condition Top-1 accuracy on DVS128 Gesture (EventMamba backbone), stratified by the recording's native illumination tag.
	The curriculum improves over the baseline under every condition, with the largest gain in the hardest one.
	Conditions are strongly imbalanced and we report $n$ accordingly; the \emph{natural} row is shown for completeness but is too small ($n=39$) to support a conclusion, and no test recordings carry the \emph{fluorescent+LED} tag
	Bold marks the better (FiLM-Rate curriculum) value per row.}
\label{tab:lighting-supp}
\small
\setlength{\tabcolsep}{6pt}
\begin{tabular}{lcccr}
\toprule
Lighting condition & Baseline & FiLM-Rate (curr.) & $\Delta$ & $n$ \\
\midrule
lab                & 55.87 & \best{65.47} & +9.60 & 12{,}905 \\
LED                & 88.71 & \best{91.38} & +2.67 & 3{,}375 \\
fluorescent        & 80.67 & \best{83.15} & +2.48 & 5{,}809 \\
\midrule
natural$^*$        & 76.92 & \best{87.18} & +10.26 & 39 \\
\bottomrule
\addlinespace[2pt]
\multicolumn{5}{l}{\footnotesize $^*$Sample count too small for a reliable
estimate; reported for completeness only.}
\end{tabular}
\end{table}

\section{Per-bucket natural-rate accuracy, full breakdown}\label{sec:s3}

\Cref{tab:bucket} expands the summary in~\cref{sec:exp:main} (``Main results''): per-bucket natural-rate accuracy of the top three methods (b0 = lowest event rate, b6 = highest).

\begin{table}[h]
\centering
\caption{Per-bucket natural-rate accuracy (\%) of the top three methods (b0 = lowest event rate, b6 = highest).
The curriculum improves over FiLM (no aug) in 6 of 7 buckets; the single regression (b6) is discussed in~\cref{sec:limits}.
Bold marks the best value per column.}
\label{tab:bucket}
\footnotesize
\setlength{\tabcolsep}{3pt}
\begin{tabular}{lccccccc|cc}
\toprule
Method & b0 & b1 & b2 & b3 & b4 & b5 & b6 & Mean & Worst \\
\midrule
Baseline                   & 65.1 & 65.9 & 68.3 & 59.8 & 60.6 & 56.6 & 50.0 & 60.9 & 50.0 \\
FiLM-Rate (no aug)         & 65.9 & 72.3 & 69.1 & 67.9 & 62.3 & 57.8 & \best{55.7} & 64.4 & \best{55.7} \\
\textbf{FiLM-Rate (curr.)} & \best{67.5} & \best{76.3} & \best{72.7} & \best{70.3} & \best{67.1} & \best{63.1} & 53.6 & \best{67.2} & 53.6 \\
$\Delta$ vs.\ baseline     & +2.4 & +10.4 & +4.4 & +10.5 & +6.5 & +6.5 & +3.6 & +6.3 & +3.6 \\
\bottomrule
\end{tabular}
\end{table}

\section{FiLM parameter analysis}\label{sec:s4}

\begin{figure}[h]
\centering
\includegraphics[width=\textwidth]{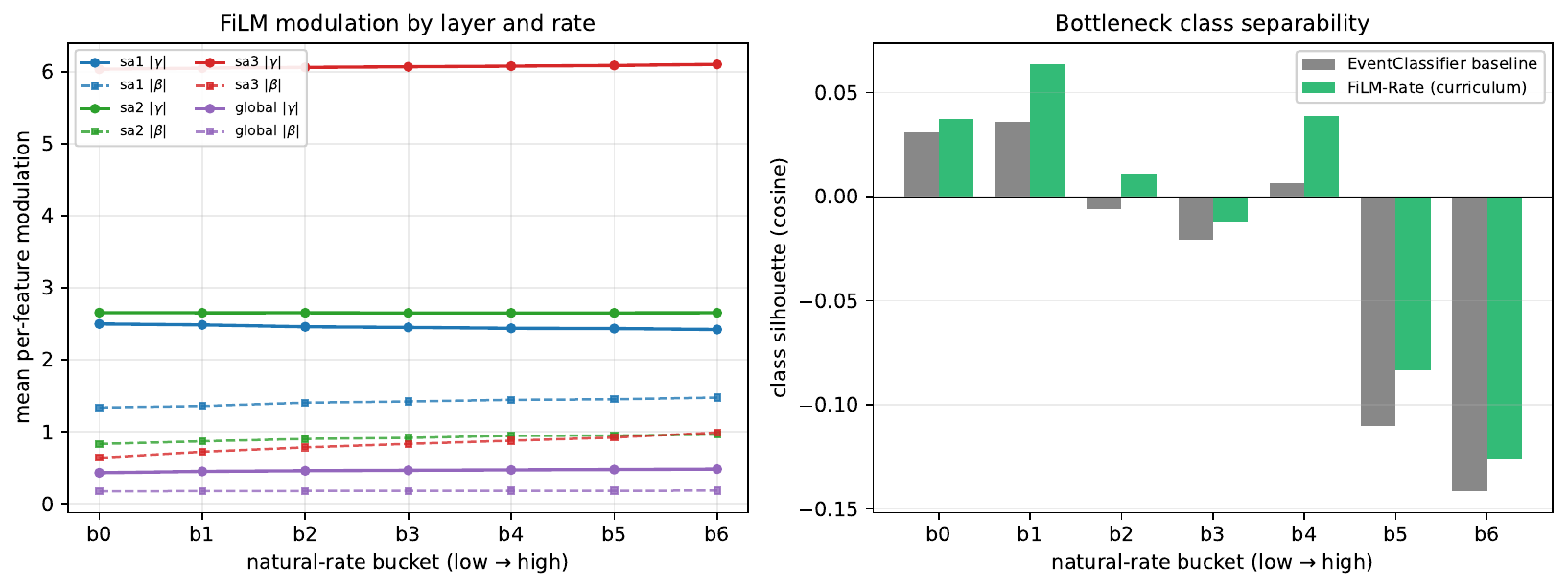}
\caption{\textbf{Left:} mean learned scale parameter, $|\gamma|$ (solid) and learned shift parameter, $|\beta|$ (dashed) per FiLM layer across natural-rate buckets.
The FiLM modulation is non-negligible at every layer and, for most channels, increases slightly with event rate.
The per-node layers (SA1--SA3) carry signal that the variant which only modulates the pooled feature representation $f_E$ (\ie{}, after global pooling) cannot represent.
\textbf{Right:} cosine-silhouette of the class structure in the pooled feature representation $f_E$, per bucket: the curriculum's features are more class-separable everywhere, with the largest gaps in the buckets where its accuracy gains are largest (\cref{tab:bucket}).}
\label{fig:film}
\end{figure}

\Cref{fig:film} expands on~\cref{sec:feature-analysis} (``What Is FiLM Actually Computing?''), visualizing the learned FiLM parameters and the resulting feature separability per natural-rate bucket.

\section{Variance decomposition of the per-recording log-rate, by dataset}\label{sec:s5}

\Cref{tab:variance-supp} expands the summary in~\cref{sec:multi}: the between-class share of the per-recording log-rate variance, against the curriculum's accuracy gain, across all four datasets studied.

\begin{table}[h]
\centering
\caption{Between-class share of the per-recording log-rate variance, and the curriculum's accuracy gain over the baseline, across four datasets.
	The gain tracks the fraction of rate variance that separates classes, not the size of the rate spread itself.
	For N-Cars, of a total log-rate variance of $0.9062$, the within-class component is $0.9060$ and the between-class component only $0.0003$.}
\label{tab:variance-supp}
\small
\setlength{\tabcolsep}{6pt}
\begin{tabular}{lcc}
\toprule
Dataset & Between-class variance share & $\Delta$ curr.\ vs.\ baseline (pp) \\
\midrule
N-Cars          & 0.0003 & $-0.5$ \\
CIFAR10-DVS     & 0.133  & $+5.2$ \\
N-Caltech101    & 0.374  & $+6.3$ \\
DVS128 Gesture  & 0.438  & $+14.0$ \\
\bottomrule
\end{tabular}
\end{table}

\end{document}